\documentclass[runningheads]{llncs}

\usepackage{graphicx}
\usepackage{amsmath}
\usepackage{amssymb}
\usepackage{booktabs}
\usepackage{makecell}
\usepackage{microtype}
\usepackage{hyperref}
\usepackage{multirow}
\usepackage{subcaption}
\usepackage{placeins}
\usepackage{float}
\usepackage[table]{xcolor}

\definecolor{fairgreen}{RGB}{220, 245, 225}
\definecolor{fairred}{RGB}{255, 225, 225}

\begin{document}

\title{Does Predictive Fairness Imply Recourse Fairness? An Empirical Study Across Models, Search and Mitigation Methods}

\titlerunning{Does Predictive Fairness Imply Recourse Fairness?}

\author{Mustafa Cavus\orcidID{0000-0002-6172-5449} \and
Yigitcan Yardimci\orcidID{0009-0006-7574-2430}}

\authorrunning{M. Cavus and Y. Yardimci}

\institute{Eskisehir Technical University, Department of Statistics, Eskisehir, T\"{u}rkiye\\
\email{mustafacavus@eskisehir.edu.tr}}

\maketitle

\begin{abstract}
Explainability and algorithmic fairness represent two fundamental pillars of trustworthy machine learning. While recent audits uncover demographic disparities in post-hoc feature attributions, the equity of actionable, recourse-oriented explanations remains largely unaddressed from an empirical perspective. In consequential financial domains such as credit risk scoring, adverse automated decisions demand equitable counterfactual explanations that minimize the social burden of recourse across privileged and unprivileged groups. In this paper, we conduct an extensive empirical study across five representative benchmark datasets, three predictive model classes, and three counterfactual generation paradigms to examine whether predictive fairness transfers to recourse fairness. Our empirical findings across extensive model and mitigation configurations reveal that predictive fairness interventions do not directly translate to recourse fairness within credit scoring systems, as boundary adjustments can operate disparately on explanation distances. Furthermore, multi-objective evolutionary optimization consistently delivers the most recourse fairness, outperforming instance-based retrieval. Finally, pre-processing via preferential boundary resampling achieves a favorable alignment between outcome and recourse fairness, whereas loss reweighting induces an operational trade-off that elevates recourse disparity.

\end{abstract}

\section{Introduction}
Predictive machine learning models are widely used to automate critical decisions across the financial sector, ranging from consumer credit scoring and loan approval to institutional distress prediction~\cite{gunonu2026explainable}. However, empirical evidence indicates that predictive models trained on historical data often inherit systemic biases, leading to disparate impacts against unprivileged groups characterized by attributes such as sex, age, or race~\cite{chouldechova2017fair,kleinberg2017inherent}. To satisfy compliance requirements established by regulatory frameworks such as the EU AI Act{\renewcommand{\thefootnote}{$\dagger$}\footnote{Regulation 2024/1689 of the European Parliament and of the Council laying down harmonised rules on artificial intelligence.}}, the research community focuses on two complementary requirements: algorithmic fairness~\cite{mehrabi2021survey} and explainable artificial intelligence~\cite{karimi2021survey}.

Within the algorithmic fairness literature, credit scoring models have received considerable attention because training instances inherently reflect systemic biases, producing trade-offs between lender profitability and demographic disparity~\cite{hurlin2022fairness,kozodoi2022fairness}. Conventional interventions predominantly target predictive fairness, seeking predictive fairness through metrics such as Demographic Parity or Equalized Odds~\cite{hardt2016equality}. Concurrently, post-hoc explainability techniques have evolved to provide rationales for complex black-box predictors. However, auditing model transparency is vulnerable to explanation manipulation~\cite{dimanov2020you} and fairwashing~\cite{aivodji2021characterizing}, since post-hoc explanations can conceal discriminatory dependencies without rectifying internal model bias. Furthermore, recent audits reveal that explanation fidelity itself exhibits systematic quality disparities across privileged and unprivileged groups~\cite{balagopalan2022road,fontana2026fairness,zhao2022fairness}. Nonetheless, these empirical audits remain confined to feature attributions, which describe historical model weights rather than providing practical assistance.

In consequential financial settings, denied applicants require actionable recourse~\cite{ustun2019actionable} to overturn unfavorable credit decisions through viable explanations~\cite{karimi2021survey}. Counterfactual explanations offer recommendations outlining what minimal adjustments an individual must perform to achieve a favorable outcome, utilizing search heuristics ranging from local instance matching to multi-objective optimization~\cite{dandl2025counterfactuals,wexler2019whatif}. Yet, achieving equality of effort and minimizing the social burden of recourse across privileged and unprivileged groups is rarely guaranteed~\cite{barrainkua2025revisiting,raimondi2022equality}. As highlighted by Artelt and Hammer~\cite{artelt2022explain}, counterfactual explanations often exhibit disparate complexity across privileged and unprivileged groups, placing unprivileged individuals under higher recourse costs. While they formulated a regularized search penalizing group complexity discrepancies during generation, their formulation requires gradient access to differentiable models and cannot evaluate tree ensembles. Furthermore, their framework does not balance multi-objective trade-offs nor examine how predictive bias mitigations interact with recourse fairness under model-agnostic, black-box settings. In this context, foundational theoretical studies formalized recourse fairness through equalized effort~\cite{gupta2019equalizing} and substantive equality of opportunity~\cite{vonkugelgen2022fairness}, demonstrating that outcome-fair classifiers can impose disparate optimal action costs across privileged and unprivileged groups~\cite{bell2025effort}. Furthermore, unstable recourse across similar individuals undermines individual fairness and plausibility~\cite{artelt2021evaluating}. Simultaneously, empirical audits in explanation fairness remain almost exclusively confined to feature attribution techniques such as SHAP and LIME, showing that explanation fidelity~\cite{balagopalan2022road} and stability~\cite{fontana2026fairness,zhao2022fairness} degrade for unprivileged groups. Consequently, an empirical gap remains in the literature examining how state-of-the-art counterfactual generators behave across privileged and unprivileged groups, and whether standard algorithmic fairness interventions alleviate or worsen recourse disparities.

In this work, we bridge these two streams through an empirical benchmark across credit risk scoring domains. By systematically auditing predictive models, counterfactual generators, and bias mitigation interventions, we investigate how algorithmic choices govern recourse disparities under real-world financial constraints. Our main contributions are summarized as follows:
\begin{enumerate}
    \item We conduct an empirical audit evaluating whether predictive model classes across linear models, bagging ensembles, and boosting algorithms systematically induce differing levels of recourse disparity across privileged and unprivileged groups.
    \item We benchmark three distinct counterfactual search paradigms, comprising instance-based exemplar matching, rule-pruned nearest instance search, and multi-objective evolutionary optimization, establishing how continuous Pareto search mitigates subgroup disparities compared to sample-dependent retrieval.
    \item We systematically audit the downstream transfer of algorithmic fairness mitigations to counterfactual quality, revealing why data-level boundary resampling harmonizes outcome and recourse fairness while loss reweighting exacerbates recourse disparities.
    \item We formulate an empirical auditing framework to quantify counterfactual recourse disparities across privileged and unprivileged groups in credit scoring, establishing directional metrics to evaluate disparate effort burdens in algorithmic decision-making.
\end{enumerate}

\section{Preliminaries}
\subsection{Predictive Fairness}
Let $\mathcal{D} = \{(\mathbf{x}_i, s_i, y_i)\}_{i=1}^n$ be a dataset where $\mathbf{x} \in \mathcal{X} \subset \mathbb{R}^d$ represents input features, $s \in \mathcal{S} = \{0, 1\}$ denotes a binary protected attribute, where $s=1$ represents the privileged group and $s=0$ the unprivileged group, and $y \in \{0, 1\}$ denotes the ground-truth outcome. A decision model $f: \mathcal{X} \to [0, 1]$ predicts $\hat{y} = \mathbb{I}(f(\mathbf{x}) \ge \tau)$ where $\tau$ is a decision cutoff, typically set to $0.5$.

Statistical Parity asserts that positive acceptance rates should be independent of the protected attribute:
\begin{equation}
P(\hat{Y} = 1 \mid S = 0) = P(\hat{Y} = 1 \mid S = 1).
\end{equation}
In practical fairness auditing, this is quantified via the Statistical Parity Ratio (SPR)~\cite{wisniewski2022fairmodels}, defined as the positive prediction rate of the unprivileged group divided by that of the privileged group. Similarly, Equal Opportunity mandates equal True Positive Rates between groups, evaluated via the Equal Opportunity Ratio (EOR)~\cite{hardt2016equality}. These ratios are defined as follows:

\begin{equation}
\text{SPR} = \frac{P(\hat{Y}=1 \mid S=0)}{P(\hat{Y}=1 \mid S=1)},
\quad
\text{EOR} = \frac{P(\hat{Y}=1 \mid Y=1,S=0)}{P(\hat{Y}=1 \mid Y=1,S=1)}.
\end{equation}

Based on the US Equal Employment Opportunity Commission Four-Fifths Rule{\renewcommand{\thefootnote}{$\ddagger$}\footnote{Uniform Guidelines on Employee Selection Procedures (1978), 29 CFR \S 1607.4(d), Adverse impact and the ``four-fifths rule.''}}, fairness metric ratios between $0.80$ and $1.25$ are considered acceptable in this study~\cite{wisniewski2022fairmodels}.

\subsection{Counterfactual Recourse}
Given an individual $\mathbf{x}$ receiving an adverse prediction $\hat{y} = 0$ with $f(\mathbf{x}) < \tau$, a counterfactual explanation $\mathbf{x}^* \in \mathcal{X}$ seeks a minimally perturbed profile that flips the model prediction to the favorable class $\hat{y}^* = 1$ with $f(\mathbf{x}^*) \ge \tau$, as formulated in previous studies~\cite{karimi2021survey}. Formally, this is framed as an optimization problem:
\begin{equation}
\mathbf{x}^* = \arg\min_{\mathbf{x}' \in \mathcal{X}} \mathcal{L}_{\text{pred}}(f(\mathbf{x}'), y_{\text{desired}}) + \lambda \, \text{dist}(\mathbf{x}, \mathbf{x}'),
\end{equation}
where $\text{dist}(\cdot, \cdot)$ denotes Gower's distance for mixed tabular data, and $\mathcal{L}_{\text{pred}}$ penalizes failure to reach the target class. Crucially, actionable recourse requires holding immutable demographic traits strictly unchanged with $\mathbf{x}^*_j = \mathbf{x}_j$ for the protected feature index $j$, ensuring recommendations alter only modifiable financial behaviors.

As established in counterfactual evaluation frameworks~\cite{dandl2025counterfactuals}, counterfactual recourse is characterized across five core properties: Proximity, measuring the distance $\text{dist}(\mathbf{x}, \mathbf{x}^*)$ reflecting the physical or financial effort required to overturn the decision; Sparsity, measuring the $\ell_0$ norm $\|\mathbf{x} - \mathbf{x}^*\|_0$ to quantify the count of modified attributes; Minimality, measuring $|f(\mathbf{x}^*) - \tau|$ indicating the decision margin above the threshold as a confidence diagnostic; Plausibility, measuring $\min_{\mathbf{x}_{\text{train}}} \text{dist}(\mathbf{x}^*, \mathbf{x}_{\text{train}})$ to ensure manifold support; and Validity, confirming that the model classifies the recommendation into the target class with $f(\mathbf{x}^*) = y_{\text{desired}}$. In our downstream disparity analysis, we focus on the actionable effort burdens of Proximity and Sparsity alongside the decision margin metric Minimality. Validity is enforced as a strict eligibility criterion. Across all evaluated model classes, search paradigms, and privileged and unprivileged groups, the generators achieve a 100 percent validity rate with zero instance exclusions, ensuring that comparative recourse disparities remain completely unskewed by sample attrition. Meanwhile, Plausibility monitors data manifold distance rather than individual action effort.

\subsection{Recourse Disparity}
\begin{figure}[H]
\centering
\includegraphics[width=0.40\textwidth]{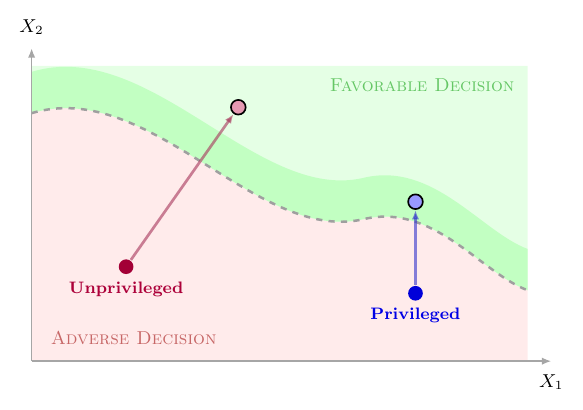}
\caption{Geometric manifestation of counterfactual recourse disparity across a non-linear credit underwriting boundary. While privileged applicants obtain low-cost recourse explanations near the boundary, unprivileged individuals must make larger changes to their features, falling into the disparate impact region and resulting in recourse disparity.}
\label{fig:recourse_fairness_illustration}
\end{figure}

Following early conceptual formulations~\cite{gupta2019equalizing,vonkugelgen2022fairness}, Recourse Fairness posits that individuals from historically unprivileged subgroups should not systematically face higher burdens to secure a favorable classification than privileged individuals. For any counterfactual effort or margin metric $M \in \{\text{Proximity}, \text{Sparsity}, \text{Minimality}\}$ where higher values correspond to greater individual effort or modification burden, we define the directional Recourse Disparity Burden ($B_M$) against the unprivileged group as:
\begin{equation}
B_M = \max\left(0,\, \mathbb{E}_{\mathbf{x} \sim \mathcal{X}_{S=0}}[M(\mathbf{x}, \mathbf{x}^*)] - \mathbb{E}_{\mathbf{x} \sim \mathcal{X}_{S=1}}[M(\mathbf{x}, \mathbf{x}^*)]\right).
\end{equation}
When recourse effort favors the unprivileged subgroup, $B_M = 0$, whereas values strictly greater than zero quantify the excess disparity burden imposed on unprivileged applicants. As illustrated in Figure~\ref{fig:recourse_fairness_illustration}, even when individuals start from adverse classification regions, non-linear decision boundaries may require unprivileged applicants to traverse substantially longer recourse trajectories than privileged peers to reach the favorable decision zone.

\section{Experiments}
\subsection{Datasets}
To ensure robust empirical evaluation across diverse decision contexts, we conduct experiments across five benchmark datasets representing consumer lending, credit scoring, and capacity-to-pay evaluations. Table~\ref{tab:dataset_summary} summarizes the sample size, feature dimensionality, protected demographic attribute, and privileged and unprivileged groups of each domain.

\vspace{0.3em}
\begin{table}[H]
\centering
\caption{Benchmark datasets, protected demographic attributes, and evaluated subgroups.}
\label{tab:dataset_summary}
\resizebox{\textwidth}{!}{
\begin{tabular}{@{}lrccp{3.0cm}p{3.0cm}@{}}
\toprule
Dataset & $n$ & $d$ & Protected & Privileged & Unprivileged \\
\midrule
German Credit~\cite{hofmann1994german} & 800 & 20 & \texttt{sex\_marital} & Male Single & Female / Non-Single \\
Taiwan Credit~\cite{yeh2009taiwan} & 7{,}000 & 23 & \texttt{SEX} & Male & Female \\
FICO HELOC~\cite{fico2018heloc} & 7{,}000 & 23 & \texttt{OldestTrade} & Mature ($\ge 180$ mos.) & Young ($< 180$ mos.) \\
Give Me Credit~\cite{kaggle2011givemecredit} & 25{,}000 & 10 & \texttt{age} & Mature ($\ge 35$ yrs.) & Young ($< 35$ yrs.) \\
Adult Income~\cite{becker1996adult} & 32{,}561 & 12 & \texttt{sex} & Male & Female \\
\bottomrule
\end{tabular}
}
\end{table}
\vspace{0.3em}

These benchmarks span diverse sample sizes with $n \in [800, 32{,}561]$, feature dimensionalities, and both demographic attributes such as sex and temporal factors including credit history maturity and age in financial decision-making. Continuous features are standardized to unit variance, and missing values are imputed prior to model estimation.

\subsection{Models and Mitigations}
For each dataset, instances are partitioned into training sets of 75 percent and independent test sets of 25 percent. We train three model classes: regularized Logistic Regression representing linear parametric baselines; Random Forest ensemble bagging classifiers configured with 100 probability trees; and Gradient Boosting Machines configured with interaction depth 3.

To examine algorithmic bias mitigation, we evaluate three preprocessing methods~\cite{wisniewski2022fairmodels}. We intentionally focus on data-level pre-processing interventions because counterfactual generation operates directly within the model's learned feature representation space, where boundary geometry directly dictates recourse trajectories. Reweighting balances instance weights in the loss function inversely proportional to subgroup outcome rates; Uniform Resampling oversamples underrepresented demographic categories to establish group frequency parity; and Preferential Resampling duplicates or discards instances situated near the decision boundary based on prediction uncertainty.

\subsection{Recourse Generation}
Counterfactual recourse is generated across three distinct algorithmic search paradigms~\cite{dandl2025counterfactuals}. The What-If approach~\cite{wexler2019whatif} conducts nearest neighbor exemplar search directly over observed positive instances in the training manifold. The NICE algorithm~\cite{burghmans2022nice} performs nearest instance counterfactual search optimizing explicitly for feature sparsity through iterative single-feature substitution. Finally, the MOC framework~\cite{dandl2020moc} employs multi-objective evolutionary optimization using NSGA-II to balance proximity, sparsity, target probability, and data manifold distance concurrently. Because MOC yields a set of non-dominated Pareto-optimal solutions, the candidate minimizing Gower feature distance to the factual instance is selected as the representative recourse explanation, directly following the principle of minimal individual effort.

For each group, we sample up to $n_{\text{eval}} = 50$ adverse decision instances from the independent test set that require algorithmic recourse to reverse an unfavorable prediction. In smaller tabular benchmarks with limited unprivileged test instances (specifically German Credit, where the total test set contains 200 instances yielding 14 and 22 adverse instances for unprivileged and privileged groups, respectively), all available adverse individuals are exhaustively evaluated. Across the five benchmark datasets, three model classes, and three explanation paradigms, this sampling procedure produces a dataset of 3,357 individual counterfactual evaluations for granular statistical and distributional analysis. The complete experimental suite, curated tabular benchmarks, pre-trained models, and replication scripts are made publicly available at \url{https://github.com/mcavs/recourse-fairness}.

\section{Results}
\subsection{Influence of Model Class on Recourse Fairness}
\begin{figure}[H]
\centering
\includegraphics[width=\textwidth]{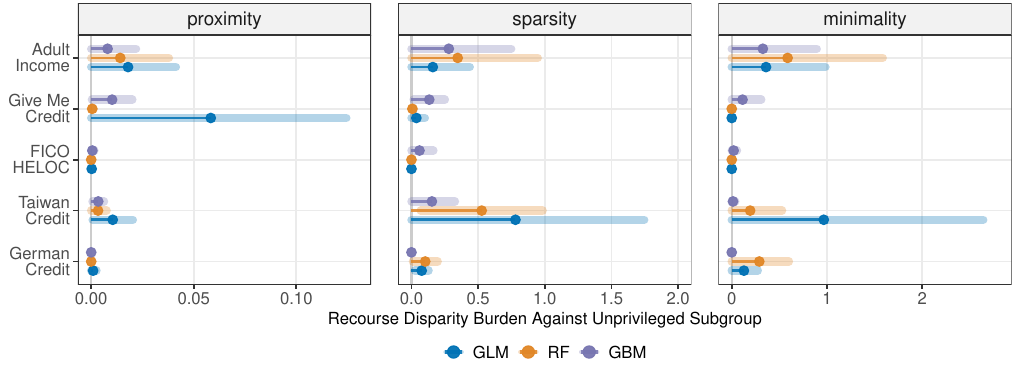}
\caption{Cross-model class comparison of counterfactual recourse disparities across predictive model classes and benchmark credit datasets. Markers indicate mean disparity with 1-SD oval variance bands relative to the perfect equity baseline.}
\label{fig:rq1_model_recourse_gaps}
\end{figure}

To evaluate whether underlying model classes systematically induce disparities in algorithmic recourse, Figure~\ref{fig:rq1_model_recourse_gaps} compares counterfactual recourse disparities across \textit{proximity} gap, \textit{sparsity} gap, and \textit{minimality} gap for canonical model families including regularized Logistic Regression, Random Forest, and Gradient Boosting Machines across five benchmark datasets.

The empirical evaluation in Figure~\ref{fig:rq1_model_recourse_gaps} demonstrates that recourse fairness disparities do not differ substantially across underlying predictive model classes. In line with our directional disparity formulation, when recourse effort favors the unprivileged subgroup, the disparity burden is evaluated as zero. Across all five benchmark domains, linear models, bagging ensembles, and gradient boosting classifiers exhibit comparable magnitudes of subgroup disparities across \textit{proximity}, \textit{sparsity}, and \textit{minimality}. While GLM exhibits higher feature \textit{sparsity} burdens in Taiwan Credit, measuring 0.78 features versus 0.15 for GBM, these variations reflect localized decision boundaries rather than systematic model biases. Furthermore, the wider 1-SD oval variance bands observed for German Credit across all model classes directly capture the empirical variance arising from its smaller sample size ($n=14$ adverse unprivileged test instances), whereas larger cohorts exhibit tightly bounded distributions.

To statistically verify whether model classes systematically differ in recourse fairness, we conduct non-parametric Friedman tests with dataset-level blocking to account for the paired experimental structure across benchmarks. The tests confirm that differences across model classes remain statistically non-significant across all evaluated recourse dimensions: \textit{proximity} ($\chi^2(2) = 5.69, p = 0.058$), \textit{sparsity} ($\chi^2(2) = 0.36, p = 0.836$), and \textit{minimality} ($\chi^2(2) = 2.48, p = 0.289$). These results establish that model complexity and inductive bias do not intrinsically govern recourse fairness disparities. Rather, as analyzed in Section~\ref{subsec:rq2_methods}, recourse inequity is predominantly driven by the optimization objectives and search constraints of the counterfactual generation algorithms themselves.

\subsection{Equity Across Counterfactual Search Paradigms}
\label{subsec:rq2_methods}
\begin{figure}[H]
\centering
\includegraphics[width=\textwidth]{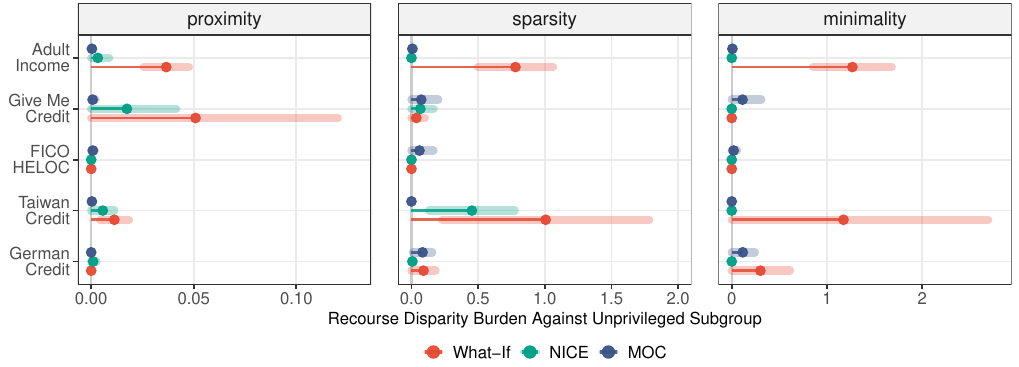}
\caption{Comparative evaluation of counterfactual recourse disparities across explanation generation paradigms and benchmark credit datasets. Lollipop markers denote group disparity means, bounded by 1-SD oval variance bands relative to the perfect equity baseline.}
\label{fig:rq2_methods_lollipop}
\end{figure}

To determine which algorithmic paradigm produces the most equitable counterfactual explanations, Figure~\ref{fig:rq2_methods_lollipop} compares the recourse fairness disparity burdens across What-If, NICE, and MOC across all benchmark datasets.

The empirical evaluation in Figure~\ref{fig:rq2_methods_lollipop} reveals substantial operational disparities across algorithmic paradigms. Multi-objective evolutionary optimization via MOC consistently minimizes recourse burdens on unprivileged groups, achieving near-zero disparity across \textit{proximity} and \textit{sparsity} across all five benchmarks. Non-parametric Friedman tests confirm statistically significant differences among counterfactual explanation paradigms in \textit{minimality} ($\chi^2(2) = 11.09, p = 0.0039$) and \textit{proximity} ($\chi^2(2) = 8.59, p = 0.0136$), with \textit{sparsity} exhibiting visible variations ($\chi^2(2) = 5.21, p = 0.0739$). Post-hoc pairwise comparisons confirm that What-If significantly exacerbates unprivileged effort burdens ($p < 0.05$), whereas MOC preserves parity without imposing differential modification costs. Because MOC navigates a continuous multi-objective optimization space independent of training sample density, it avoids penalizing data-sparse unprivileged subgroups. Multi-objective evolutionary search thus emerges as the most equitable algorithmic paradigm for actionable recourse.

\subsection{Downstream Transfer of Algorithmic Fairness Mitigations}
\begin{figure}[H]
\centering
\includegraphics[width=\textwidth]{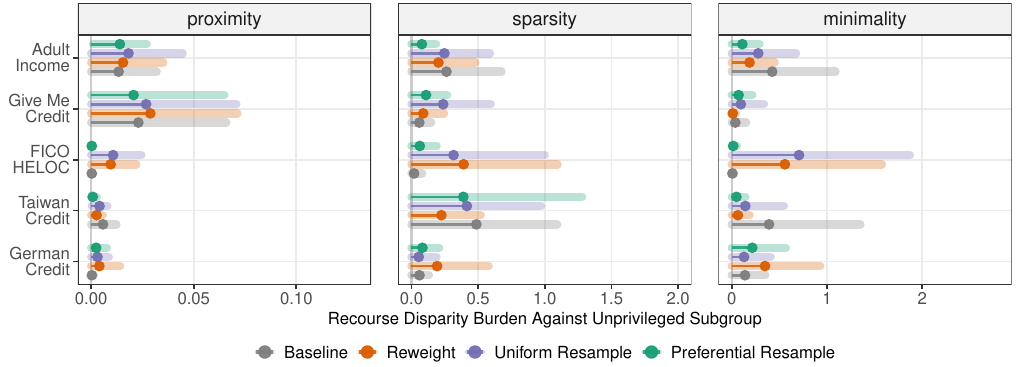}
\caption{Impact of algorithmic fairness mitigation strategies on counterfactual recourse disparities across benchmark credit datasets.}
\label{fig:rq3_mitigation_gaps}
\end{figure}

To empirically evaluate whether predictive fairness guarantees recourse fairness, we first document the predictive fairness achieved by each bias mitigation strategy in Table~\ref{tab:predictive_fairness_full}, evaluating the Statistical Parity Ratio and Equal Opportunity Ratio across all predictive model classes and benchmark datasets before and after intervention. We then trace the downstream consequences of these interventions on counterfactual recourse disparity burdens in Figure~\ref{fig:rq3_mitigation_gaps}, comparing Reweight, Uniform Resampling, and Preferential Resampling against unmitigated baselines across all benchmark domains.

\vspace{-0.8em}
\begin{table}[H]
\centering
\caption{Predictive fairness metrics across model classes, benchmark datasets, and bias mitigation strategies. Green cell background indicates fair outcomes satisfying the four-fifths criterion between 0.80 and 1.25, while red indicates unfair outcomes falling outside this range.}
\label{tab:predictive_fairness_full}
\resizebox{\linewidth}{!}{
\begin{tabular}{llcccccccc}
\toprule
& & \multicolumn{4}{c}{Statistical Parity Ratio} & \multicolumn{4}{c}{Equal Opportunity Ratio} \\
\cmidrule(lr){3-6} \cmidrule(lr){7-10}
Model & Dataset & Baseline & Reweight & Uniform & Preferential & Baseline & Reweight & Uniform & Preferential \\
\midrule
\multirow{5}{*}{GLM}   & German Credit   & \cellcolor{fairgreen}0.967 & \cellcolor{fairgreen}1.006 & \cellcolor{fairgreen}1.023 & \cellcolor{fairgreen}1.141 & \cellcolor{fairgreen}0.992 & \cellcolor{fairgreen}0.992 & \cellcolor{fairgreen}1.017 & \cellcolor{fairgreen}1.094 \\
                       & Taiwan Credit   & \cellcolor{fairred}0.570 & \cellcolor{fairred}0.776 & \cellcolor{fairred}0.752 & \cellcolor{fairgreen}1.144 & \cellcolor{fairred}0.727 & \cellcolor{fairgreen}0.906 & \cellcolor{fairgreen}0.929 & \cellcolor{fairred}1.355 \\
                       & FICO HELOC      & \cellcolor{fairred}1.777 & \cellcolor{fairgreen}1.227 & \cellcolor{fairgreen}1.223 & \cellcolor{fairgreen}1.185 & \cellcolor{fairred}1.331 & \cellcolor{fairgreen}1.000 & \cellcolor{fairgreen}0.994 & \cellcolor{fairgreen}0.979 \\
                       & Give Me Credit  & \cellcolor{fairgreen}0.998 & \cellcolor{fairgreen}0.999 & \cellcolor{fairgreen}1.000 & \cellcolor{fairgreen}1.001 & \cellcolor{fairgreen}0.998 & \cellcolor{fairgreen}0.998 & \cellcolor{fairgreen}0.999 & \cellcolor{fairgreen}0.999 \\
                       & Adult Income    & \cellcolor{fairred}0.320 & \cellcolor{fairred}0.632 & \cellcolor{fairred}0.599 & \cellcolor{fairgreen}1.120 & \cellcolor{fairgreen}0.925 & \cellcolor{fairred}1.307 & \cellcolor{fairred}1.290 & \cellcolor{fairred}1.773 \\
\midrule
\multirow{5}{*}{RF}    & German Credit   & \cellcolor{fairgreen}0.995 & \cellcolor{fairgreen}1.046 & \cellcolor{fairgreen}1.083 & \cellcolor{fairgreen}1.017 & \cellcolor{fairgreen}0.998 & \cellcolor{fairgreen}1.017 & \cellcolor{fairgreen}1.046 & \cellcolor{fairgreen}1.012 \\
                       & Taiwan Credit   & \cellcolor{fairred}0.690 & \cellcolor{fairred}0.671 & \cellcolor{fairred}0.774 & \cellcolor{fairgreen}0.821 & \cellcolor{fairred}0.767 & \cellcolor{fairred}0.745 & \cellcolor{fairgreen}0.870 & \cellcolor{fairgreen}0.860 \\
                       & FICO HELOC      & \cellcolor{fairred}1.672 & \cellcolor{fairred}1.255 & \cellcolor{fairred}1.400 & \cellcolor{fairred}1.323 & \cellcolor{fairgreen}1.221 & \cellcolor{fairgreen}1.004 & \cellcolor{fairgreen}1.100 & \cellcolor{fairgreen}1.039 \\
                       & Give Me Credit  & \cellcolor{fairgreen}0.963 & \cellcolor{fairgreen}0.984 & \cellcolor{fairgreen}0.976 & \cellcolor{fairgreen}0.985 & \cellcolor{fairgreen}0.983 & \cellcolor{fairgreen}0.992 & \cellcolor{fairgreen}0.991 & \cellcolor{fairgreen}0.992 \\
                       & Adult Income    & \cellcolor{fairred}0.330 & \cellcolor{fairred}0.557 & \cellcolor{fairred}0.480 & \cellcolor{fairgreen}0.924 & \cellcolor{fairgreen}0.912 & \cellcolor{fairgreen}1.249 & \cellcolor{fairgreen}1.088 & \cellcolor{fairred}1.614 \\
\midrule
\multirow{5}{*}{GBM}   & German Credit   & \cellcolor{fairgreen}0.857 & \cellcolor{fairgreen}0.973 & \cellcolor{fairgreen}0.929 & \cellcolor{fairgreen}1.025 & \cellcolor{fairgreen}0.884 & \cellcolor{fairgreen}0.998 & \cellcolor{fairgreen}0.944 & \cellcolor{fairgreen}1.012 \\
                       & Taiwan Credit   & \cellcolor{fairred}0.721 & \cellcolor{fairred}0.771 & \cellcolor{fairred}0.778 & \cellcolor{fairgreen}0.877 & \cellcolor{fairgreen}0.842 & \cellcolor{fairgreen}0.940 & \cellcolor{fairgreen}0.944 & \cellcolor{fairgreen}1.153 \\
                       & FICO HELOC      & \cellcolor{fairred}1.654 & \cellcolor{fairgreen}1.115 & \cellcolor{fairgreen}1.219 & \cellcolor{fairgreen}1.115 & \cellcolor{fairgreen}1.247 & \cellcolor{fairgreen}0.931 & \cellcolor{fairgreen}0.976 & \cellcolor{fairgreen}0.931 \\
                       & Give Me Credit  & \cellcolor{fairgreen}0.975 & \cellcolor{fairgreen}1.003 & \cellcolor{fairgreen}1.002 & \cellcolor{fairgreen}1.011 & \cellcolor{fairgreen}0.987 & \cellcolor{fairgreen}0.997 & \cellcolor{fairgreen}0.996 & \cellcolor{fairgreen}1.013 \\
                       & Adult Income    & \cellcolor{fairred}0.326 & \cellcolor{fairred}0.560 & \cellcolor{fairred}0.527 & \cellcolor{fairgreen}1.069 & \cellcolor{fairgreen}0.924 & \cellcolor{fairred}1.262 & \cellcolor{fairgreen}1.249 & \cellcolor{fairred}1.701 \\
\bottomrule
\end{tabular}
}
\end{table}

The empirical results in Figure~\ref{fig:rq3_mitigation_gaps} and Table~\ref{tab:predictive_fairness_full} across five benchmark domains and all three explanation paradigms reveal that pre-processing mitigations affect predictive fairness, yet predictive fairness gains do not inherently translate to recourse fairness. As documented in Table~\ref{tab:predictive_fairness_full}, interventions such as loss reweighting and uniform resampling successfully eliminate predictive disparities in specific domains; for example, in FICO HELOC, reweighting restores GLM EOR from an unfair 1.331 to a compliant 1.000, and in Taiwan Credit, it elevates GLM EOR from 0.727 to 0.906. However, as evidenced in Figure~\ref{fig:rq3_mitigation_gaps}, this statistical predictive fairness fails to guarantee recourse fairness: reweighting and uniform resampling elevate the mean \textit{proximity} gap to 0.0121 and 0.0125, respectively, and the \textit{minimality} gap to 0.234 and 0.270, exceeding unmitigated baseline levels of 0.0076 and 0.200. This difference suggests that in Table~\ref{tab:predictive_fairness_full}: while reweighting enforces demographic acceptance parity by penalizing majority misclassifications, it does not alter empirical sample density in historically underrepresented regions. While direct geometrical measurement of high-dimensional decision boundaries remains outside our empirical scope, we hypothesize that reweighting shifts the decision threshold toward sparser feature spaces for unprivileged applicants, satisfying aggregate demographic parity while increasing the perturbation distance required for individual recourse.

In contrast, preferential boundary resampling achieves better alignment between outcome and recourse fairness. It reduces demographic disparities across challenging benchmarks in Table~\ref{tab:predictive_fairness_full}, transforming adult income SPR from low baseline values of 0.320--0.330 into compliant ratios of 0.924--1.120, and Taiwan Credit SPR up to 1.144, while concurrently minimizing recourse burdens in Figure~\ref{fig:rq3_mitigation_gaps}, reducing the mean \textit{minimality} gap to 0.0925 and lowering the \textit{sparsity} gap to 0.144. Non-parametric Friedman tests across mitigation strategies confirm a statistically significant difference in \textit{proximity} gap ($\chi^2(3) = 15.29, p = 0.0016$), with \textit{sparsity} ($\chi^2(3) = 1.40, p = 0.706$) and \textit{minimality} ($\chi^2(3) = 1.96, p = 0.580$) exhibiting consistent performance. We hypothesize that by selectively reinforcing unprivileged instances in uncertain boundary regions, preferential boundary resampling enhances local sample density along decision contours, thereby reducing disparate perturbation requirements for unprivileged individuals. Data-level boundary resampling thus emerges as an effective paradigm for improving both predictive and recourse fairness.

\section{Conclusion}
In this paper, we investigated the fairness of counterfactual explanations across predictive model classes and generation paradigms within tabular credit scoring domains. Our findings challenge the premise that predictive fairness guarantees recourse fairness, demonstrating that predictive fairness interventions do not necessarily alleviate recourse disparities, as loss adjustments can operate disparately on explanations in credit scoring. Furthermore, our results establish multi-objective evolutionary counterfactuals as the most equitable explanation paradigm, and identify preferential boundary resampling as a promising intervention for improving both predictive and recourse fairness.

\textit{Limitations.} This study focuses on conventional bias mitigation methods. Including more recently proposed methods could provide additional insights into predictive and recourse fairness.

\textit{Future directions.} Developing intrinsic fairness-constrained counterfactual algorithms that simultaneously optimize multiple objectives, such as proximity, sparsity, and plausibility under group parity constraints, without penalizing privileged individuals, remains an open research problem. In addition, future work involves geometrically auditing decision boundary shifts under loss reweighting via local margin sampling, extending these evaluations to causal recourse graphs that account for actionable intervention dependencies, auditing explanation stability and burden sharing under predictive multiplicity in Rashomon sets~\cite{ganesh2026beyond}, and examining the temporal dynamics of recourse fairness in longitudinal credit environments. Finally, extending these evaluations to broader families of fairness-aware data preparation mitigations~\cite{figueiredo2026scikit} represents a promising avenue for understanding how emerging preprocessing interventions affect recourse fairness.

\bibliographystyle{splncs04_unsrt}
\bibliography{references}

\end{document}